\pdfoutput=1

\documentclass[11pt]{article}

\usepackage{ACL2023}

\usepackage{times}
\usepackage{latexsym}
\usepackage{hyperref}
\usepackage{url}
\usepackage{amsmath}
\usepackage{graphicx}
\usepackage{xspace}
\usepackage{booktabs}
\usepackage{tabularx}
\usepackage{amssymb}   
\usepackage{amsfonts}  
\usepackage{amsthm} 
\usepackage{makecell}
\usepackage{multirow}
\usepackage{subcaption}
\usepackage{wrapfig}
\usepackage{booktabs}
\usepackage{amsmath}
\usepackage{bbm}
\usepackage{multirow}
\usepackage{array}
\usepackage{float}
\usepackage{wrapfig}
\usepackage{enumitem}
\usepackage{longtable}

\usepackage{hyperref}
\usepackage{listings}
\usepackage{url}
\usepackage{varwidth}

\def\memory{DPM\xspace}
\def\model{PlugLM\xspace}
\newcommand{\todo}[1]{\textcolor{red}{\textbf{[TODO]}}}

\newcommand\rct{\textsc{RCT}\xspace}

\newcommand\chemprot{\textsc{ChemProt}\xspace}
\newcommand\arccite{\textsc{ACL-ARC}\xspace}
\newcommand\sciie{\textsc{SciERC}\xspace}
\newcommand\hp{\textsc{HyperPartisan}\xspace}

\newcommand\ag{\textsc{AGNews}\xspace}
\newcommand\helpful{\textsc{Helpfulness}\xspace}
\newcommand\imdb{\textsc{IMDB}\xspace}

\newcommand\news{\textsc{News}\xspace}
\newcommand\med{\textsc{BioMed}\xspace}
\newcommand\cs{\textsc{CS}\xspace}
\newcommand\realnews{\textsc{RealNews}\xspace}
\newcommand\reviews{\textsc{Reviews}\xspace}

\newcommand\amazon{\textsc{Amazon} reviews\xspace}
\newcommand\gorc{\textsc{S2ORC}\xspace}

\usepackage[T1]{fontenc}

\usepackage[utf8]{inputenc}

\usepackage{microtype}

\usepackage{inconsolata}

\title{\mbox{Decouple knowledge from parameters for plug-and-play language modeling}}

\newcommand\icst{$^1$}
\newcommand\ruc{$^2$}
\newcommand\kaust{$^3$}
\newcommand\bigai{$^4$}
\newcommand\gai{$^5$}

\author{
Xin Cheng \icst, 
Yankai Lin \ruc, 
Xiuying Chen \kaust, 
Dongyan Zhao \icst$^{,}$\bigai$^{,}$\gai$^{,}$\thanks{\;\;Corresponding author.},  
Rui Yan \ruc $^*$ \\
\icst~Wangxuan Institute of Computer Technology, Peking University\\
\ruc~Gaoling School of Artificial Intelligence, Renmin University of China\\
\kaust~Computational Bioscience Reseach Center, KAUST \quad \bigai BIGAI, Beijing, China \\
\gai National Key Laboratory of General Artificial Intelligence\\
{\tt chengxin1998@stu.pku.edu.cn}\quad {\tt yankailin@ruc.edu.cn} \\
{\tt xiuying.chen@kaust.edu.sa} \quad {\tt zhaody@pku.edu.cn} \\
{\tt ruiyan@ruc.edu.cn}\\
}

\begin{document}
\maketitle
\begin{abstract}
Pre-trained language models~(PLM) have made impressive results in various NLP tasks. It has been revealed that one of the key factors to their success is the parameters of these models implicitly learn all kinds of knowledge during pre-training.
However, encoding knowledge implicitly in the model parameters has two fundamental drawbacks. First, the knowledge is neither editable nor scalable once the model is trained, which is especially problematic in that knowledge is consistently evolving. Second, it lacks interpretability and prevents humans from understanding which knowledge PLM requires for a certain problem. In this paper, we introduce \model, a pre-training model with differentiable plug-in memory~(DPM). The key intuition is to decouple the knowledge storage from model parameters with an editable and scalable key-value memory and leverage knowledge in an explainable manner by knowledge retrieval in the \memory. 
To justify this design choice, we conduct evaluations in three settings including:~(1) domain adaptation. \model obtains 3.95 F1 improvements across four domains on average without any in-domain pre-training. (2) knowledge update. \model could absorb new knowledge in a training-free way after pre-training is done. (3) in-task knowledge learning. \model could be further improved by incorporating training samples into \memory with knowledge prompting\footnote{Code available at \url{https://github.com/Hannibal046/PlugLM}}.

\end{abstract}

\section{Introduction}
Large pre-trained language models~(PLM)~\citep{peters-etal-2018-deep,devlin-etal-2019-bert,radford2018improving} have become a revolutionary breakthrough in NLP area. Optimized by carefully designed self-supervised objectives on unlabeled corpus and fine-tuned on downstream tasks, PLMs perform remarkably well in a wide range of NLP benchmarks. Recent studies~\citep{warstadt2019investigating,petroni2019language} have revealed that one of the key factors to the success of PLMs is that the parameters of these models implicitly learn various types of knowledge in the pre-training corpus. Owing to these learned syntactic, semantic, factual and commonsense knowledge, PLMs show great understanding, generalization and reasoning abilities in multiple downstream tasks~\citep{rogers2020primer,izacard2022few}. As \citet{geva2021transformer} pointed out, the feed-forward layers~(FFN), constituting two-thirds of a transformer model's parameters, are essentially key-value memories and store all kinds of knowledge of PLM. The first linear layer of FFN acts like a set of sparsely activated keys detecting input patterns while the second is the corresponding value. To aggressively capture more knowledge, larger PLMs are continuously proposed, from 110M BERT~\citep{devlin-etal-2019-bert} to 530B MT-NLG~\citep{smith2022using}, yet PLM has not reached upper bound~\citep{ouyang2022training}. 

However, a fundamental question still remains: \textbf{For PLM, is it the optimal way to implicitly encode knowledge in its parameters? } We argue that the implicit knowledge encoding approach has two fundamental drawbacks. 
First, the learned knowledge is neither editable nor scalable once the model is trained~(e.g., BERT doesn't know what is a BERT). Nevertheless, world knowledge is actually infinite and evolving. We thus would never expect an ever-large model to capture all the knowledge in its parameters and to be continuously re-trained for the newly coming one. 
Second, the current PLMs lack interpretability at the knowledge level. Implicit knowledge encoding fails to provide provenance for model’s prediction and makes PLM a black box preventing humans from understanding which knowledge PLM requires for a certain problem. 

\begin{figure*}[t]
		\centering
		\begin{subfigure}{\textwidth}
            \centering
            \includegraphics[width=\linewidth]{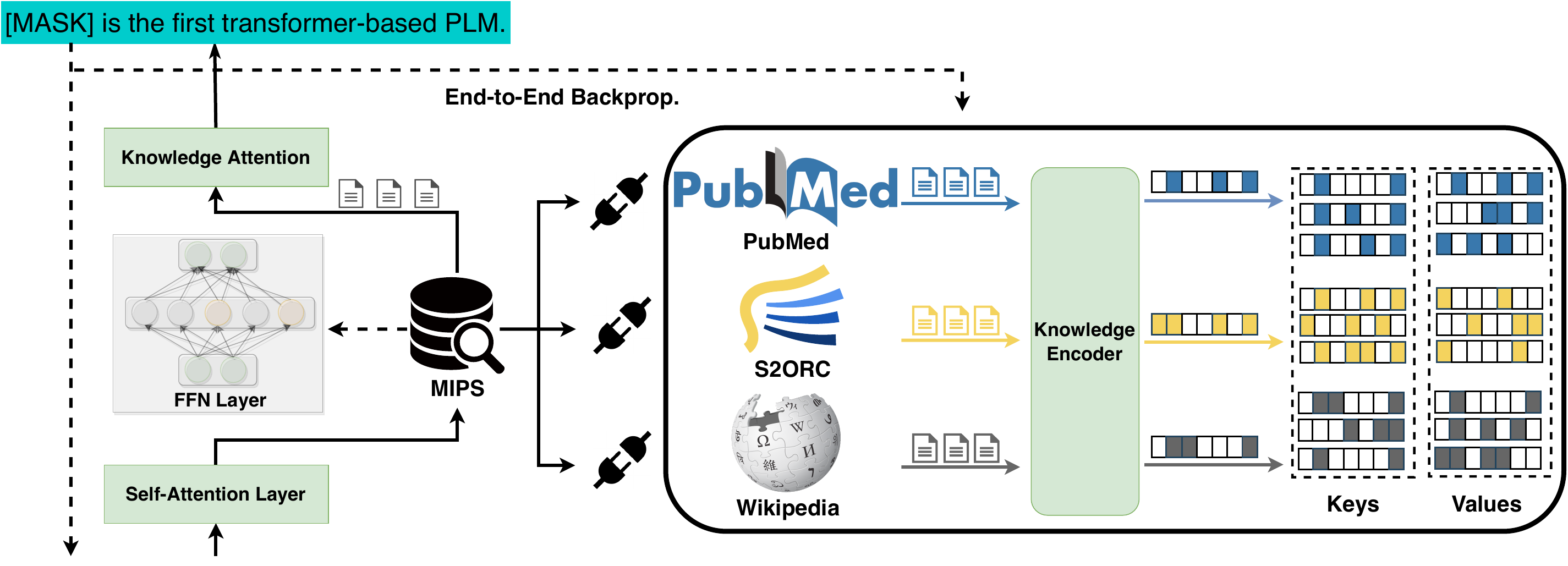}
        \end{subfigure}%
		\caption{Overview of our \model. We replace FFN in PLM with a Differentiable Plug-in key-value Memory~(DPM) by which PLM could store and leverage knowledge in an explainable manner.}
		\label{fig:model}
\end{figure*}

In this work, we propose a novel architecture of PLM, \model, which decouples the knowledge storage from model parameters and explicitly leverages the knowledge in an explainable manner. As shown in Figure~\ref{fig:model}, we balance the functionality of FFN layer with a differentiable plug-in key-value memory~(\memory), which is highly scalable as well as editable. Each slot of \memory encodes the knowledge to a pair of key and value, and thus we can explicitly retrieve the required knowledge in natural language from \memory rather than unnamed vectors in FFN.

To justify the design choice of decoupling the knowledge from parameters, we conduct extensive evaluations under different settings. 
In the domain adaptation setting, \model could be easily adapted to different domains with pluggable in-domain memory---obtaining 3.95 F1 improvements across four domains on average and up to 11.55 F1 improvement on \arccite citation intent classification dataset, without any in-domain pre-training. 
In the knowledge update setting, \model could absorb new knowledge after pre-training is done in a training-free way by knowledge updating operation in the \memory, with an improvement up to 4 F1 scores in \texttt{LINNAEUS} NER dataset. 
\model could further be improved by incorporating training samples into \memory with knowledge prompting as a kind of in-task knowledge.

\section{Related Work}
\paragraph{Investigating FFN} 
Feed-forward layers constitute two-thirds of a transformer model's parameters and are essential to unveil modern PLMs~\citep{geva2021transformer,geva2022transformer}. A surge of works have investigated the knowledge captured by FFN~\citep{dai2022knowledge,meng2022locating,geva2021transformer,geva2022transformer,jiang2020can,yao2022kformer,wallat2020bertnesia}. Based on the view that FFN is essentially an unnormalized key-value memory network, \citet{dai2022knowledge} detects knowledge neurons in FFN and edit specific factual knowledge without fine-tuning. \citet{meng2022locating} modifies FFN weights to update specific factual associations using Rank-One Model Editing. \citet{yao2022kformer} injects knowledge into the FFN via BM25. \citet{dai2022neural} and \citet{lample2019large} enhance the model by expanding the size of FFN with extra trainable keys and values. 

\paragraph{Knowledge-Augmented Language Model} There are two lines of works to equip PLM with knowledge. 
The first is introduce additional Knowledge Graph~(KG) and knowledge-based training signal~(e.g., entity linking) into the language model pre-training, like ERNIE~\citep{zhang2019ernie,sun2019ernie}, KnowBERT~\citep{peters2019knowledge} and KEPLER~\citep{wang2021kepler}. Another line of works adopt retrieval mechanism to incorporate knowledge, either symbolic~\citep{verga2021adaptable,agarwal2021knowledge,fevry2020entities} or texual~\citep{guu2020retrieval,lewis2020retrieval,borgeaud2022improving,lewis2020pre,verga2021adaptable,de2021mention}. They formulate the task as retrieve then predict process by using extra neural dense retriever or sparse retriever to find most relevant supporting knowledge and combine it with input using either concatenation~\citep{guu2020retrieval,lewis2020retrieval}, attention methods~\citep{de2021mention,chen2022augmenting} or interpolation~\citep{khandelwal2019generalization,DBLP:journals/corr/abs-2205-12674}

\model differs from previous works in that we do not try to equip the model with additional knowledge to perform knowledge-intensive tasks. The key insight is to transform FFN architecture into deep retrieval in the interest of decoupling the knowledge which would otherwise be stored in the parameters and this is orthogonal to all retrieval-augmented PLMs.

\section{Preliminary}
\paragraph{Feed-forward Layers} Transformer~\citep{vaswani2017attention}, the backbone for all PLMs, is made of stacked self-attention~(Self-Attn) and feed-forward~(FFN) layers. The former captures the contextual interaction among inputs and the latter process each input independently.  Let $x \in \mathbb{R}^{d_{1}}$ be a vector as input, the FFN could be formulated as:
\begin{equation}
\label{equation:ffn}
	\text{FFN}(x) = \sigma(x\cdot \mathbf{W_{1}^\top})\cdot \mathbf{W_{2}}
\end{equation}
where $\mathbf{W_{1}},\mathbf{W_{2}}\in \mathbb{R}^{d_2\times d_1}$ and $\sigma$ is the activation function. The bias term is omitted for brevity.

\paragraph{Key-Value Memory Network}
The Key-Value Memory Network~\citep{weston2014memory,sukhbaatar2015end} corresponds to $d_2$ key-value pairs and each key/value is a vector in $\mathbb{R}^{d_1}$. They are the generalization of the way knowledge is stored~\citep{eric2017key,miller2016key}. For an input $x \in \mathbb{R}^{d_{1}}$, there are two stages for a key-value memory network. First, the lookup~(addressing) stage would compute the matching degree between $x$ and each key. In the second stage, $x$ would be transformed by the weighted sum of values according to the distribution of the matching degree in the first stage. We can formally define it as:
\begin{equation}
\label{memorynetwork}
	\text{MemoryNetwork}(x) = \text{softmax}(x\cdot \mathbf{K^\top})\cdot \mathbf{V}
\end{equation}
where $\mathbf{K},\mathbf{V}\in \mathbb{R}^{d_2\times d_1}$. Comparing equation (\ref{equation:ffn}) and (\ref{memorynetwork}), we could find that the FFN is an unnormalized version of MemoryNetwork. The keys in FFN are pattern detectors and would be activated only when certain patterns occur in the input. This explains how FFN stores knowledge in a key-value manner~\citep{geva2021transformer,sukhbaatar2019augmenting}. 

\section{\model}
The overall architecture of \model is illustrated in Figure~\ref{fig:model}. Because FFN is essentially a key-value memory network~\citep{geva2021transformer,dai2022knowledge,meng2022locating}, \model creatively decouples the knowledge storage from model parameters by replacing\footnote{Because different layers in transformer capture different knowledge, the lower layer for shallow patterns while the upper layers for more semantic ones~\citep{geva2021transformer,jawahar-etal-2019-bert}, we only consider replacing FFN in Top-L layers with \memory while keeping FFN in the lower layers untouched to encode the intrinsic language understanding knowledge as detailed in \S\ref{sec:tuning_model}.} FFN with a Differential Plug-in key-value Memory, DPM~(\S\ref{section:dpm_construction}) and conducting knowledge retrieval in DPM with knowledge attention~(\S\ref{section:memory_fusion}) for explicit knowledge usage instead of storing all knowledge implicitly in the model parameters. In \S\ref{section:training}, we detailedly explain how \model is trained in both pre-training and fine-tuning stages. 

\subsection{Differential Plug-in Memory}  
\label{section:dpm_construction}
In this paper, we view n-th knowledge $d_{n} = \{t^1_{n},t^2_{n},...,t^{|d_{n}|}_{n}\}$ as consecutive tokens from unlabeled corpora as in \citet{guu2020retrieval}. For each $d_n$, we get its dense representation $h_n$ from a knowledge encoder $\text{KnowEncoder}(\cdot)$:
 \begin{equation}
	h_n = \text{AttnPooling}(\text{E}_{\text{Token}}(d_n)+\text{E}_{\text{Pos}}(d_n))
\end{equation}
where AttentivePooling function~\citep{xu2021fusing,cheng2023personalized} corresponds to a trainable pattern detector aggregating information from a sequence of input. And $\text{E}_{\text{Token}}$ and $\text{E}_{\text{Pos}}$ denote token embedding and positional embedding. Then we use two independent mapping functions to project $h_n$ to the key space and value space:
\begin{align}
	\quad k_n = \mathbf{W_k} \cdot h_n+\mathbf{b_k} \\	
	v_n = \mathbf{W_v}\cdot h_n  +\mathbf{v_k}
\end{align}
where $\mathbf{W_k}$, $\mathbf{W_v}$, $\mathbf{b_k}$ and $\mathbf{v_k}$ are trainable parameters. And DPM is a triplet of $\langle \mathbb{D},\mathbb{K},\mathbb{V} \rangle$:
\begin{align}
	\mathbb{D} &= \{d_1,d_2,...,d_{|\mathbb{D}|}\} \\
	\mathbb{K} &= \{k_1,k_2,...,k_{|\mathbb{D}|}\}\\
	\mathbb{V} &= \{v_1,v_2,...,v_{|\mathbb{D}|}\}	
\end{align}


\subsection{Memory Fusion} 
\label{section:memory_fusion}
For hidden states $h\in {\mathbb{R}^{l\times d}}$ from Self-Attn, FFN would transform $h$ with unnormalized key-value memory as in Equation~(\ref{equation:ffn}). Our key insight is that instead of interacting with unnamed vectors in FFN, we conduct Maximum Inner Product Search~(MIPS) to retrieve knowledge in natural language from $\langle \mathbb{D},\mathbb{K},\mathbb{V} \rangle$ where each triplet corresponds to one knowledge along with its key and value representation. For $h$, we first get its sentence-level representation $z$ by an attentive pooling function $z=\text{AttentivePooling}(h)$, then we use $z$ as the query vector to $\langle \mathbb{D},\mathbb{K},\mathbb{V} \rangle$. Since PLM is internally sparse~\citep{li2022large},
 we only consider Top-N knowledge $\mathbb{D}_{z}$ with corresponding keys $\mathbb{K}_{z}$ and values $\mathbb{V}_{z}$:
\begin{align}
	\mathbb{K}_{z} &= \text{Top-N}(\text{MIPS}(z,\mathbb{K})) \\
	\mathbb{V}_{z} &= \{v_{i}\ \text{if}\ k_{i}\ \text{in}\ \mathbb{K}_{z}\} \\
	\mathbb{D}_{z}&= \{d_{i}\ \text{if}\ k_{i}\ \text{in}\ \mathbb{K}_{z}\}
\end{align}
where Top-N also corresponds to the indexing operation. With $\mathbb{K}_{z}$ and $\mathbb{V}_{z}$, we use knowledge attention to fuse retrieved knowledge into our model:
\begin{align}
 	\text{Attention}(h,\!\mathbb{K}_{z},\!\mathbb{V}_{z}\!)\!=\!\text{softmax}(\!\frac{h\mathbb{K}_{z}^\top}{\sqrt{d}}\!)\mathbb{V}_{z}
\end{align}
where $d$ is the head dimension.
By knowledge retrieval and fusion, we explore an interpretable way to incorporate knowledge into the model where $\mathbb{D}_{z}$ is the actual knowledge that PLM would leverage. And direct modification on $\mathbb{D}$ without changing model parameters empowers \model with much flexibility and scalability in domain adaptation~(\S\ref{sec:domain_adaptation}) and knowledge update  ~(\S\ref{sec:knowledge_update}) scenarios.

\subsection{Training}
\label{section:training}
The backbone of our model is a multi-layer bidirectional transformer encoder~\citep{devlin-etal-2019-bert}. There are two phases in our framework: pre-training and fine-tuning. In the pre-training phase, to make the whole training process end-to-end trainable, we use asynchronous index refreshing to optimize our model as done in \citet{guu2020retrieval} and \citet{cai2021neural}. Concretely, we update the indices of DPM every T steps. The MIPS results are based on the stale index while the scores of selected Top-N results are recomputed using $\text{KnowEncoder}(\cdot)$ which facilitates the gradient flow back to memory. The training objective is Masked Language Modeling~\citep{devlin-etal-2019-bert} where we randomly mask tokens in a sentence and ask \model to predict it. In the pre-training phase, Wikipedia is chosen as the source of knowledge and in the domain adaptation fine-tuning stage, corpora from other domains are treated as knowledge sources detailed in $\S$\ref{sec:domain_adaptation}.  More details are shown in Appendix~\ref{appendix:pretraining_details}. In the fine-tuning phase, the $\mathbb{K}$ and $\mathbb{V}$ of DPM are fixed, and we view it as an editable and scalable knowledge lookup table.

\section{Experiments}
\model mainly tries to decouple the knowledge storage from parameters and leverage knowledge in an explainable way. We conduct comprehensive experiments to show the superiority of this novel architecture: we could easily adapt the model to different domains without in-domain pre-training by switching DPM~(\S\ref{section:domain_adaptive_post_training} and \S\ref{section:in_domain_pre_training}), alleviate catastrophic forgetting by storing DPM~(\S\ref{section:domain_adaptive_post_training}), inject new knowledge into the model by enlarging DPM~(\S\ref{sec:knowledge_update}), further enhance the model by injecting in-task knowledge into DPM~(\S\ref{sec:example_base_learning}) and unveil the black-box PLM with direct access to the knowledge retrieved from DPM~(Appendix~\ref{appendix:case_study}). We also carefully examine each key design in \model and point the direction for future work in \S\ref{sec:tuning_model}.

\subsection{Domain Adaptation}
\label{sec:domain_adaptation}
Learning robust and transferable representation has been the core of language model pre-training~\citep{peters2019knowledge}. For the general-purposed PLM to generalize well on domain-specific tasks, endowing the model with domain knowledge via in-domain training remains the go-to approach~\citep{gururangan2020don,whang2020effective,zhang-etal-2020-multi-stage,10.1145/3543507.3583389}.
In this section, we show that without any in-domain pre-training, \model could flexibly adapt to multiple domains with domain-specific \memory. For the existing PLM encoding knowledge in parameters, this is a challenging task in that it can not guarantee the generalization across multiple domains due to catastrophic forgetting~\citep{kirkpatrick2017overcoming} and sometimes it is even computationally unaffordable to keep training the super large models~\citep{smith2022using,brown2020language}. 

We consider two adaptation scenarios: domain adaptive post-training~(\S\ref{section:domain_adaptive_post_training}) and in-domain pre-training~(\S\ref{section:in_domain_pre_training}). The former is conducted after PLM was trained on the general domain and the latter trains a domain-specific PLM from scratch.

\begin{table*}[t]
    \centering
    \resizebox{0.95\linewidth}{!}{
    \begin{tabular}
    {ccccccccccc}
    \toprule[1.2pt]
 	\textbf{Model} & 
 		\multicolumn{2}{c}{\textbf{\underline{\med}}} & 
 		\multicolumn{2}{c}{\textbf{\underline{\cs}}} &
 		\multicolumn{2}{c}{\textbf{\underline{\news}}} &
 		\multicolumn{2}{c}{\textbf{\underline{\reviews}}}  &
 		\\[0.5ex]

        & CHEM. & RCT &ACL. &SCI. & HYP. & AG. & HP. & IMDB & \thead{Avg.\\Gain} & \thead{Avg. \\ Cost}\\
        \midrule[0.5pt]
        WikiBERT & 77.72 & 86.52 & 61.58 & 79.95 & 83.54 & 93.38 & 67.62 & 89.79 & - & -\\
        \hspace{0.5em} + DAPT  & 78.24 & 86.71 & 67.56 & 80.82 & 86.22 & 93.49 & 68.11 & 90.12 & +1.40 & 47.7 h\\
        \hspace{0.5em} $\lnot$ DAPT & 75.82 & 86.11 & 62.11 & 78.42 & 80.12 & 93.31 & 68.11 & 89.54 & -0.82 & -\\
        \hspace{0.5em} + DACT & 76.34 & 86.11 & 61.19 & 78.56 & 80.52 & 93.29 & 68.08 & 89.88 & -0.77 & -\\
        \midrule[0.5pt]
		REALM & 78.28 & 85.12 & 62.07 & 78.41 & 84.12 & 92.58 &  67.06 & 90.56 & - & -\\
		\hspace{0.5em} + DAA & 79.32 & 85.98 & 68.92 & 80.41 & 85.36 & 92.61 & 68.51 & \textbf{93.01} & +1.98 & \underline{6.3 h}\\
		\hspace{0.5em} $\lnot$ DAA & 77.61 & 85.12 &  64.78 & 75.31 & 82.28 & 92.41 & 66.13 & 91.21 & -0.41 & -\\
		\hspace{0.5em} + DAR & 80.56 & 85.32 & 70.12 & 81.16 & 86.58 & 93.01 & 67.42 & 92.16 &+2.26 & \underline{6.3 h}\\
        \midrule[0.5pt]
        \model & 78.02 & 87.12 & 63.77 & 78.56 & 84.32 & 93.23 & 67.83 & 91.24 & - & -\\
        \hspace{0.5em} + DAA & \underline{82.56} & \underline{88.13} & \underline{72.51} & \textbf{83.00} & \underline{88.16} & \textbf{94.11} & \underline{69.28} & 92.56 & \underline{+3.28} & \textbf{0.16 h}\\
		\hspace{0.5em} $\lnot$ DAA & 77.98 & 86.13 & 64.78 & 78.13 & 84.18 & 92.99 & 67.56 & 90.88 & -0.18 & -\\
        \hspace{0.5em} + DAR & \textbf{83.80 }& \textbf{88.98} & \textbf{75.32} & \underline{82.56} & \textbf{89.26} & \underline{93.55} & \textbf{69.41} & \underline{92.78} & \textbf{+3.95} & \textbf{0.16 h}\\
        \addlinespace[0.15em]
        \bottomrule[1.2pt]
    \end{tabular}}
	\caption{Performance of domain adaptive post-training. Each result is averaged with five different random seeds. Reported results are test macro-F1, except for \rct  and \chemprot, for which we report micro-F1, following \citet{beltagy2019scibert}. The best scores are in bold, and the second best are underlined. 
    }
    \label{tab:dapt_result}
\end{table*}
\subsubsection{Domain Adaptive Post-Training}
\label{section:domain_adaptive_post_training}

\paragraph{Experimental Setting} Following \citet{gururangan2020don}, we conduct experiments on four domains: \med, \cs, \news and \reviews across eight domain-specific downstream tasks, in both low and high resource settings. More details can be found in Appendix~\ref{appendix:dapt_data}. When fine-tuning, we pass the final [CLS] representation to a task-specific head as in \citet{devlin-etal-2019-bert}.

We have the following baselines:~\textbf{WikiBERT} uses the architecture of $\text{BERT}_{base}$~\citep{devlin-etal-2019-bert} and is pre-trained on Wikipedia. To adapt WikiBERT to other domains, we use DAPT following the training setting in \citet{gururangan2019variational}. \textbf{REALM}~\citep{guu2020retrieval} and ~\textbf{\model} are models that have an external knowledge base and can be simply adapted to other domains with a different base. We have two adaptation strategies: \underline{DAA}, short for Domain Adaptive Addition, appends domain knowledge to the knowledge base, and \underline{DAR}, Domain Adaptive Replacement, replaces general knowledge with domain-specific knowledge in the knowledge base.

We also include the results of \underline{$\lnot$DAPT}, \underline{$\lnot$DAA} and \underline{DACT}. The former two use irrelevant domain corpora for post-training and knowledge base construction, which are used to test the robustness of the adaptation method and rule out the factor that improvements might be attributed simply to exposure to more data\footnote{Following \citet{gururangan2020don}, we use the following irrelevant domain mapping: for \news, we use a \cs LM; for \reviews, a \med LM; for \cs, a  \news LM; for \med, a \reviews LM.}. 
For DACT, Domain Adaptive Continual Training, we sequentially use DAPT for WikiBERT in multiple domains in the hope that it can capture and store knowledge from various domains in a lifelong learning way~\citep{rostami2021lifelong}.

\begin{figure*}[thbp]
		\centering
		\begin{subfigure}{\textwidth}
            \centering
    \includegraphics[width=0.9\linewidth,height=0.28\linewidth]{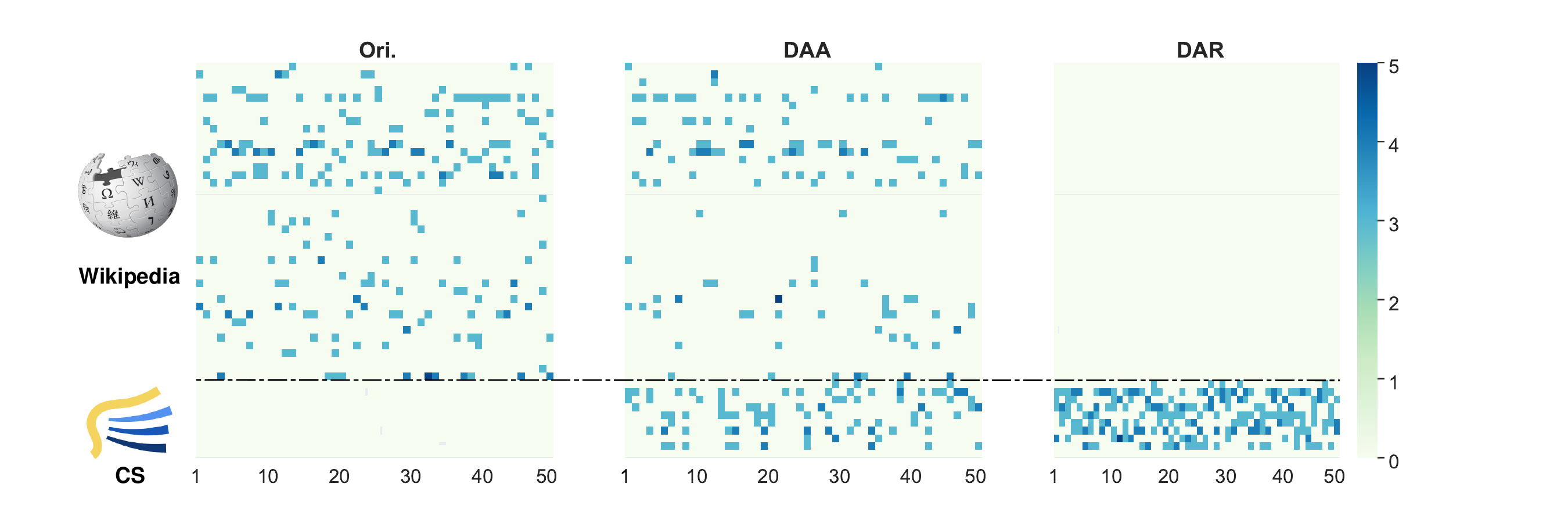}
        \end{subfigure}%
		\caption{Knowledge retrieval visualization. We randomly sample 50 samples from \arccite test set and check what kind of knowledge does \model use to solve CS-specific tasks. Each column is one sample and the row is the index of retrieved knowledge in \memory. Their corresponding F1 scores are 63.77, 72.51 and 75.32.}
		\label{fig:knowledge_distribution}
\end{figure*}

\paragraph{Experimental Results} The results are shown in Table~\ref{tab:dapt_result}. The Avg.Cost is the cost for adaptation measured by hour. For WikiBERT, it's the time to post-train  model in domain-specific corpus. For REALM and \model, it is the time to encode domain knowledge into the knowledge base. We can observe: (1)~In-domain training helps model better generalize to tasks requiring domain knowledge while irrelevant knowledge misleads the model and causes performance degradation. And by comparing $\lnot$DAPT and $\lnot$DAA, it shows that models with external knowledge base~(\model and REALM) are more robust when faced with noisy out-of-domain knowledge. (2)~For the model that implicitly encodes knowledge in the parameters, it fails to generalize across domains as the result of DACT indicates. For example, we keep training WikiBERT in \news domain after DAPT in \cs domain and fine-tune it on the \cs downstream tasks. It performs on par with model that is never exposed to \cs domain~($\lnot$DAPT). \model could alleviate this catastrophic forgetting problem by storing all kinds of knowledge in \memory and using it in a plug-and-play manner. (3)~Direct modification on external memory helps \model efficiently and effectively adapt to different domains without in-domain training. In 254$\times$ less time compared with DAPT and in 40$\times$ less time compared with REALM, \model significantly outperforms DAPT and REALM-based methods.

To further understand \model, in Figure~\ref{fig:knowledge_distribution}, we present a visualization for the distribution of actual retrieved knowledge for DAA, DAR and original \model. A clear pattern here is that with more domain knowledge involved, the model performs better~(63.77, 72.51 and 75.32) and remarkably, although pre-trained on the general domain, the \model has managed to learn what to retrieve when there are both general knowledge and domain-specific knowledge in \memory shown in DAA visualization.

\subsubsection{In-domain Pre-Training}
\label{section:in_domain_pre_training}

In-domain pre-training is another line of work for domain-specific PLM training from scratch like BioBERT~\citep{Lee2019BioBERTAP}, SciBERT~\citep{beltagy2019scibert} and FinBERT~\citep{araci2019finbert}. 
\paragraph{Experimental Setting} In this section, we choose the biomedical domain and compare \model with model in the architecture of $\text{BERT}_{base}$, pre-trained on the general domain, Wikipedia~(i.e., WikiBERT) and pre-trained on the biomedical domain, Pubmed~(i.e., PubmedBERT). The statistics of datasets and pre-training details are listed in Appendix~\ref{appendix:pretraining_data}.
We test two kinds of abilities of these PLMs. First, we test how they perform in biomed-relevant downstream tasks. Specifically, we conduct experiments on eight representative biomedical NER datasets which aim at recognizing domain-specific proper nouns in the biomedical corpus. Then we test their general language understanding ability in \texttt{GLUE}~\citep{wang2018glue} and \texttt{SQUAD}~\citep{rajpurkar2016squad,rajpurkar2018know}. For \texttt{SQUAD} and \texttt{GLUE}, the \memory is constructed from Wikipedia, and for biomedical NER, \memory is from PubMed~\citep{canese2013pubmed}.
\paragraph{Experimental Results} The results are shown in Table~\ref{tab:ner_result}.  Both pre-trained on the Wikipedia, \model outperforms WikiBERT in 8/8 NER tasks with average 1.75 F1 scores by simply switching the knowledge domain of \memory. \model also gives comparable results with PubmedBERT in \texttt{BC4CHEMD}, \texttt{JNLPBA} and \texttt{LINNAEUS} datasets. Although PubmedBERT works well for biomedical tasks, it shows less general language understanding ability and underperforms WikiBERT and \model in \texttt{GLUE}~(Table~\ref{tab:glue_result}) and \texttt{SQUAD}~(Table~\ref{tab:squad_result}), especially in low resource scenario~(i.e., \texttt{RTE}, \texttt{COLA} and \texttt{MRPC} datasets). With \memory, \model shows great flexibility and performs well in both general domain and biomedical domain. In Appendix~\ref{appendix:case_study}, we give concrete cases of \model with respect to the retrieved knowledge.
\begin{table}[thbp]
\centering
\resizebox{0.95\linewidth}{!}{
\begin{tabular}
    {ccccccc}
    \toprule[1.2pt]
    & \multicolumn{2}{c}{\bf PubmedBERT} 
          & \multicolumn{2}{c}{\bf WikiBERT} &        \multicolumn{2}{c}{\bf \model} \\[0.5ex]
& EM       & F1          & EM       & F1    & EM         & F1    \\ 		        \midrule[0.5pt]		
SQUAD(v1) &76.68 & 84.56 & \underline{81.32}    & \underline{88.68}  & \textbf{82.19}      & \textbf{89.44} \\
SQUAD(v2) &68.44 & 71.12 & \underline{72.64}    & \underline{75.89}  & \textbf{73.76}      & \textbf{76.90} \\
        \bottomrule[1.2pt]
\end{tabular}}
\caption{\texttt{SQUAD} results measured by EM and F1.}
\label{tab:squad_result}
\end{table}

\begin{table*}[htbp]
\centering	
\resizebox{0.85\linewidth}{!}{
\begin{tabular}
    {llcccc}
\toprule[1.2pt]
   Type& Dataset  & \# Annotation & WikiBERT & \model & PubmedBERT \\
\midrule
\multirow{2}{*}{Disease}       & \texttt{NCBI-disease} & 6811          & 83.65    & \underline{85.96} & \bf 88.39     \\
& \texttt{BC5CDR}       & 12694         & 80.37    & \underline{82.10}  & \bf 83.89  \\
	\midrule[0.5pt]	
\addlinespace[0.15em]
\multirow{2}{*}{Drug/Chem.}    & \texttt{BC4CHEMD}     & 79842         & 87.07    & \bf 89.93  & \underline{89.35}      \\
& \texttt{BC5CDR}       & 15411         & 88.79    & \underline{90.56}  & \bf 92.75\\	\midrule[0.5pt]	
\addlinespace[0.15em]
\multirow{2}{*}{Gene/Protein.} & \texttt{B2CGM}        & 20703         & 80.63    & \underline{82.14}  & \bf 83.16      \\
& \texttt{JNLPBA}       & 35460         & 75.49    & \bf 76.39  & \underline{76.25}      \\	\midrule[0.5pt]	
\addlinespace[0.15em]
\multirow{2}{*}{Species}       & \texttt{LINNAEUS}     & 4077          & 85.32    & \bf 87.01 & \underline{86.11}       \\
& \texttt{SPECIES-800}  & 3708          & 68.54    & \underline{69.73} & \bf 71.32   \\  
        \addlinespace[0.15em]
        \bottomrule[1.2pt]
\end{tabular}}
\caption{Performance of biomedical NER measured by F1 score across eight datasets.}
\label{tab:ner_result}
\end{table*}

\begin{table*}[htbp]
\centering
\resizebox{0.95\linewidth}{!}{
\begin{tabular}
    {lclcccccccc}
    \toprule[1.2pt]
	&\textbf{\#Paras} &\thead{\textbf{Avg.}\\\textbf{Latency}} & \textbf{RTE}   & \textbf{COLA}  & \textbf{MRPC}  & \textbf{STS-B} & \textbf{SST-2} & \textbf{QNLI}  & \textbf{QQP}   & \thead{\textbf{MNLI}\\\textbf{-(m/mm)}}\\
	\midrule[0.5pt]	
PubmedBERT &110M& $\times$1.00 &61.17 & 50.06 & 84.56 & 85.73 & 88.64 & 90.11 & 88.78 & 82.14/82.56\\        \addlinespace[0.25em]		        	
WikiBERT &110M& $\times$1.00 &\underline{65.70} & \textbf{53.53} & \underline{88.85} & \underline{88.64} & \textbf{92.32} & \underline{90.66} & \underline{89.71} & \underline{83.91/84.10} \\
	\addlinespace[0.25em]
\model& 109M &$\times$2.54&\textbf{70.40}&\underline{52.68}&\textbf{91.54}&\textbf{89.20}&\underline{91.86}&\textbf{91.28}&\textbf{90.56}&\textbf{84.56/85.35}\\
        \bottomrule[1.2pt]
\end{tabular}}
\caption{\texttt{GLUE} results. Detailed metrics and latency of each model is in Appendix~\ref{appendix:latency}}
\label{tab:glue_result}
\end{table*}

\subsection{Knowledge Update} 
\label{sec:knowledge_update}
Since the world is not fixed as a snapshot once the pre-training corpus is collected, the current PLM, no matter how large it is, fails to adapt to this changing world. For colossal PLMs like GPT-3~\citep{brown2020language} and MT-NLG~\citep{smith2022using}, efficiently fine-tuning for downstream tasks remains an open challenge, let alone re-training it on the newly coming knowledge.

\paragraph{Experimental Setting} In this section, we show that \model can efficiently absorb new knowledge by updating the $\langle \mathbb{D},\mathbb{K},\mathbb{V} \rangle$ without re-training. We consider the following two settings. (1) We only pre-train \model with limited data and gradually enlarge the \memory with unseen knowledge when fine-tuning. (2) We pre-train \model with full general-domain data and ask the model to perform domain adaptation in DAR manner by gradually increasing domain knowledge in $\langle \mathbb{D},\mathbb{K},\mathbb{V} \rangle$.

\begin{figure*}[t]
		\centering
		\begin{subfigure}{0.3\textwidth}
            \centering
            \includegraphics[width=0.9\linewidth,height=0.9\linewidth]{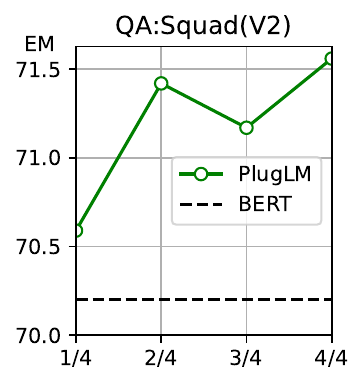}
            \caption{QA results.}
            \label{fig:knowledge_upadte_qa}
        \end{subfigure}%
		\begin{subfigure}{0.3\textwidth}
            \centering
            \includegraphics[width=0.9\linewidth,height=0.9\linewidth]{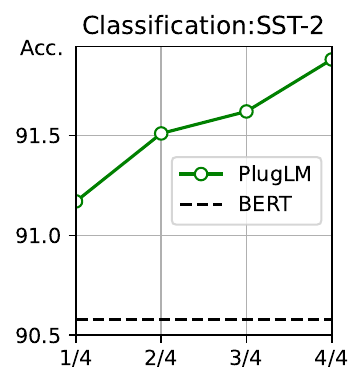}
            \caption{Classification results.}
            \label{fig:knowledge_upadte_cls}
        \end{subfigure}%
		\begin{subfigure}{0.3\textwidth}
            \centering
            \includegraphics[width=0.9\linewidth,height=0.9\linewidth]{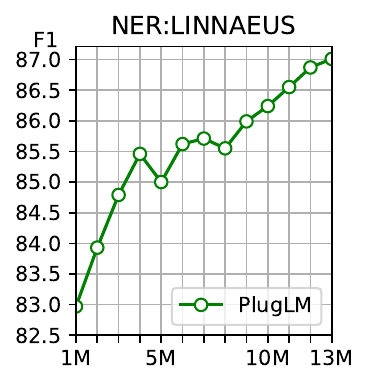}
            \caption{NER in adaptation setting.}
            \label{fig:knowledge_upadte_ner}
        \end{subfigure}%
		\caption{Knowledge update results in QA, Sentiment Classification and NER. 
        }
		\label{fig:knowledge_update}
\end{figure*}

\paragraph{Experimental Results} The results are shown in Figure~\ref{fig:knowledge_upadte_qa} and \ref{fig:knowledge_upadte_cls}. For the first setting, we test on QA~(\texttt{SQUAD}) and Sentiment Classification tasks~(\texttt{SST-2}). Both WikiBERT and \model are pre-trained with only 1/4 Wikipedia corpus. We have the following observations: 
(1)~\model trained with limited data already outperforms WikiBERT in both tasks~(0.39 EM in QA and 0.59 Accuracy in classification) which verifies the effectiveness of \model in low-resource setting; 
(2) A consistent pattern across two tasks verifies \model could absorb new knowledge simply by adding more slots in  $\langle \mathbb{D},\mathbb{K},\mathbb{V} \rangle$ without heavy re-training.

For the second setting, Figure~\ref{fig:knowledge_upadte_ner} shows our model can absorb new cross-domain knowledge under adaptation setting. It achieves a higher F1 score on the \texttt{LINNAEUS} NER dataset with increasingly more biomed-specific knowledge injected.

\subsection{In-task Knowledge}
\label{sec:example_base_learning}
Inspired by in-context learning~\citep{brown2020language} and example-augmented generation~\citep{cheng2022neural,cheng2023lift}, the training samples can also be viewed as a kind of in-task knowledge. In this section, we broaden the scope of \memory knowledge by including the training samples. 

\paragraph{Experimental Setting} Since the knowledge from Wikipedia is a textual description from domain experts while the training sample from a Question-answering NLI dataset is in the form of [Q, A, Label], this surface form distribution shift may affect the knowledge retrieval. We consider the following injection methods. (1)~Concate. We directly concatenate each training sample as a long string in the form of ``Q [SEP] A [SEP] Label" and append this to \memory. (2)~Tagged. To build the connection between model inputs and \memory, we tag each training sample by prepending a special token~([Tagged]), and use these tagged samples in both \memory and as model input. 
(3)~Knowledge Prompting. Inspired by prompting method~\citep{liu2021pre,schick2021exploiting}, we transfer in-task knowledge to knowledge in the form of Wikipedia by a natural language prompting. For example, in \texttt{QNLI} dataset, we transform [Q, A, Label] with the following prompting: ``The first sentence (doesn't) entail(s) with the second. The first sentence is [Q] and the second is [A]". We choose moderate-sized \texttt{QNLI} and \texttt{QQP} tasks because in-task knowledge injection doesn't apply to low-resource setting in our preliminary experiments. 
\paragraph{Experimental Results} The result is shown in Table~\ref{tab:in-task_knowledge}.  We can observe that \model has managed to learn from in-task knowledge and the surface-form of knowledge affect the model performance. Concatenation of training sample fails to inform \model the actual in-task knowledge~(zero retrieval in \texttt{QNLI}) and building connection between data and knowledge by a special tagged token only gives minor improvements. Instead, a well-designed knowledge prompting can help \model learn task-specific knowledge.

\begin{table}[htbp]
	\centering
	\resizebox{0.95\linewidth}{!}{
	\begin{tabular}{cccccc}
    \toprule[1.2pt]
    \bf Task &  \bf Ori. & \bf Concate. &\bf Tagged. &\bf Prompting.\\
    \midrule[0.5pt]
	QNLI & 91.28 & 91.28 & 91.37 & \bf 91.58 \\
	QQP  & 90.56 & 90.12 & 90.76 & \bf 91.47 \\
	\bottomrule[1.2pt]
    \end{tabular}}
    \caption{
    Performance of in-task knowledge on \texttt{QNLI} and \texttt{QQP} measured by accuracy.}
    \label{tab:in-task_knowledge}
\end{table}

\subsection{Tuning \model}
\label{sec:tuning_model}
\begin{figure}[thbp]
	\includegraphics[width=\linewidth,height=0.45\linewidth]{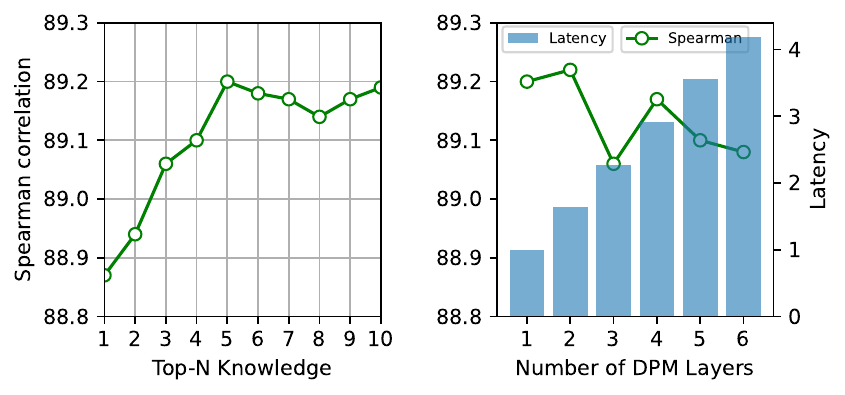}
        \caption{Effect of the number of retrieved knowledge and the number of DPM-enhanced layers in \texttt{STS-B} measured by spearman correlation.}
        \label{fig:num_knowledge}	
\end{figure}

We investigate how each key design affects the performance of \model. 
(1)~\textbf{Number of Retrieved Knowledge.} Figure~\ref{fig:num_knowledge} shows the effects of different N in STS-B dataset and the sparsely activated Top-5 knowledge proves to be optimal. 
(2)~\textbf{Layers equipped with DPM.} Considering that the  upper layers in PLM capture more semantic information~\citep{geva2021transformer}, we equip the last encoder layer with DPM in \model.  Figure~\ref{fig:num_knowledge} shows that increasing DPM-enhanced encoder layer gives minor improvements but brings much latency because of extra MIPS search. 
(3)~\textbf{FFN and \memory.} To further explore the relation between FFN and \memory, we propose two model variants. First, we replace FFN in all encoder layers with a shared DPM denoted as \model$_{\text{All}}$. Then we fuse FFN and DPM by modifying the model architecture from $\text{LayerNorm}(h+\text{KnowAttn}(h,\mathbb{K}_{h'},\mathbb{V}_{h'}))$ to $\text{LayerNorm}(h+\text{KnowAttn}(h,\mathbb{K}_{h'},\mathbb{V}_{h'})+\text{FFN}(h))$ and we name it \model$_{\text{Fuse}}$. The Spearman correlation~(more results are shown in Appendix~\ref{appendix:tuning_model}) in STS-B dataset for WikiBERT, \model$_{\text{All}}$, \model and \model$_{\text{Fuse}}$ is 88.64, 86.82, 89.20 and 89.10. We could find that \model$_{\text{All}}$, where there is no FFN, underperforms WikiBERT. And \model performs comparably with \model$_{\text{Fuse}}$. We conjecture that FFN in different layers may play different roles, which is also reported in \citet{geva2021transformer}. For the upper layer which captures more semantic knowledge~\citep{jawahar2019does}, \memory is a flexible and extensible substitution of FFN, but for lower layers, shallow features should be captured in the model parameters.

\section{Conclusion}

For the first time, we challenge the current implicit knowledge encoding mechanism for PLMs with two fundamental drawbacks and insightfully propose to decouple knowledge storage from model parameters with an editable and scalable key-value memory. Inspired by the findings that FFN stores all kinds of knowledge and is essentially a key-value memory network, we transform FFN architecture into deep retrieval with a differentiable plug-in memory~(DPM), which makes the knowledge encoding of PLMs more flexible and interpretable. Extensive experimental results in different scenarios including domain adaptation, knowledge update and in-task knowledge learning verify the design choice of \model. We believe this architectural design would pave a new direction for future research on PLM, especially for super-large PLM.

\section*{Limitations}
We discuss the limitations of \model as follows:

(1)~Despite the strong performance achieved by our approach with \memory, it results in a reduced inference efficiency at the same time due to the MIPS search. For example, \model is about two times slower than pure transformer-based models in \texttt{GLUE}. This would be more crucial when the external memory is much larger. Potential solutions to this issue include (1) constructing the memory using a coarser granularity~\citep{borgeaud2022improving}; (2) compressing DPM by semantic clustering as in \citet{tay2022transformer} or knowledge summarization as in \citet{xu2022evidence}. 

(2)~In this paper, we choose Wikipedia for \memory construction and \model pre-training. While Wikipedia is the most commonly used data source for language model pre-training~\citep{devlin-etal-2019-bert,Liu2019RoBERTaAR}, there are also many other types of knowledge not covered in Wikipedia, and how to integrate different types of knowledge~(e.g., factual, commonsense, syntactic and semantic knowledge) into our framework remains under-explored.

(3)~Although this paper proposes a general architecture that is applicable to PLMs of all kinds and sizes including bidirectional~\citep{devlin-etal-2019-bert,Liu2019RoBERTaAR,yang2019xlnet}, unidirectional~\citep{radford2018improving,radford2019language,brown2020language} and encoder-decoder-based PLM~\citep{lewis2020bart,raffel2020exploring,song2019mass}, we only experiment with bidirectional models in moderate size. In particular, we believe this architectural design would be greatly beneficial for LLM~\citep{smith2022using,chowdhery2022palm,ouyang2022training} for the following reasons: (1) the parameters of LLM could not be easily updated once the pre-training is done due to the unaffordable training cost. (2) the additional latency cost by MIPS retrieval is negligible compared with that of the whole LLM.

\bibliography{custom}
\bibliographystyle{acl_natbib}

\appendix
\clearpage
%

\section{\model Pretraining Details}
The details of \model pre-training is shown in Table~\ref{table:pluglm_adaptive_pretraining_hyperparameters}
\label{appendix:pretraining_details}
\begin{table}[h!]
    \centering
    \resizebox{0.9\linewidth}{!}{
    
    \begin{tabular}{cc}
        \toprule
        \textbf{Hyperparameter} & \textbf{Assignment}  \\
        \midrule
        vocab size & 30522 \\
        num layers with DPM & top-1 \\
        top-N & 5 \\
        number of layers & 12 \\
        attention head & 12 \\
        mlm masking & static \\
        mlm masking rate & 0.15 \\
        ffn size & 3072 \\
        max knowledge length & 288 \\
        Uncased & True \\
        memory size & 14802866 \\
        batch size & 64 \\
        gradient accumulation steps & 128 \\
        max train steps & 8000 \\
        optimizer &  FusedLAMBAMP \\
        learning rate & 1e-4 \\
        index refreshing step & 200\\
        learning rate scheduler & PolyWarmUpScheduler \\
        Warmup proportion & 0.2843 \\
        weight decay & 0.01 \\
        \bottomrule
    \end{tabular}}
    
    \caption{Hyperparameters for \model pretraining.} 
    \label{table:pluglm_adaptive_pretraining_hyperparameters}
\end{table}

\section{Data for Domain Adaptive Post-Training}
\label{appendix:dapt_data}
The detailed statistics of domain corpora for post-training is listed in the Table~\ref{tab:dapt_pretrain_data} and downstream tasks in Table~\ref{tab:dapt_down_stream_task}.

\begin{table*}[h]
\centering
\small
\begin{tabular}{m{3cm}m{7cm}rrrr}
\toprule
\bf Domain & \bf Pretraining Corpus & \bf \# Tokens &\bf Size \\
\midrule
\med
& 1.24M  papers from  \gorc \citep{lo2020s2orc}  
& 2.67B 
& 12GB \\
\cs 
& 5.07M papers from \gorc \citep{lo2020s2orc} 
& 4.3B  
& 18GB \\

\news
& 11.90M articles from \realnews \cite{zellers2019defending}
& 6.66B  
& 39GB\\
\reviews
& 24.75M \amazon \citep{he2016ups} 
& 2.11B 
& 11GB \\
\bottomrule
\end{tabular}
\caption{List of the domain-specific unlabeled datasets.}
\label{tab:dapt_pretrain_data}
\end{table*}

\begin{table*}[h]
\resizebox{0.95\linewidth}{!}{
\begin{tabular}{lllrrrrr}
\toprule
\textbf{Domain}
& \textbf{Task}            
& \textbf{Label Type} 
& \textbf{Train (Lab.)} 
& \textbf{Dev.} 
& \textbf{Test} 
& \textbf{Classes}      \\
\midrule
\multirow{2}{*}{\med} 
& \chemprot 
& relation classification
& 4169  
& 2427  
& 3469 
&       13    \\
& $^\dagger$\rct            
& abstract sent. roles 
& 18040    
& 30212 
& 30135 
& 5 \\
\midrule[0.03em]
\multirow{2}{*}{\cs}
& \arccite  
& citation intent 
& 1688          
&  114 
& 139 
& 6                     \\
& \sciie     
&  relation classification     
& 3219 
& 455  
& 974        
& 7    \\
\midrule[0.03em]
\multirow{2}{*}{\news}
& \hp 
& partisanship 
& 515 
& 65
& 65 
& 2             \\
& $^\dagger$\ag       
& topic  
& 115000    
& 5000   
& 7600  
& 4      \\
\midrule[0.03em]
\multirow{2}{*}{\reviews} 
& $^\dagger$\helpful
& review helpfulness 
& 115251 
& 5000 
& 25000  
& 2    \\
& $^\dagger$\imdb     
& review sentiment  
& 20000 
& 5000 
& 25000 
& 2       \\ 
\bottomrule
\end{tabular}
}

\caption{Specifications of the various target task datasets.
$\dagger$ indicates high-resource settings.
Sources: \chemprot \cite{Kringelum2016ChemProt30AG}, \rct \cite{dernoncourt2017pubmed}, \arccite \cite{Jurgens2018MeasuringTE}, \sciie \cite{luan2018multi}, \hp \cite{stein:2019h}, \ag \cite{zhang2015character}, \helpful \cite{mcauley2015image}, \imdb \cite{maas2011learning}. 
}
\label{tab:dapt_down_stream_task}
\end{table*}

\section{Latency}
In Table~\ref{tab:pretrain_data}, we show the detailed latency of WikiBERT and \model.
\label{appendix:latency}
\begin{table*}[htbp]
\centering
\resizebox{0.95\linewidth}{!}{
\begin{tabular}
    {
    ccccccccc
    }
    \toprule[1.5pt]
 & \textbf{RTE}   & \textbf{COLA}  & \textbf{MRPC}  & \textbf{STS-B} & \textbf{SST-2} & \textbf{QNLI}  & \textbf{QQP}   & \textbf{MNLI-(m/mm)} \\
\addlinespace[0.25em]	
 Size & 0.27K & 1.04K & 0.41K & 1.5K & 0.87K & 5.47K & 40.43K & 9.82K/9.83K \\	
 Metrics & Accuracy & Matthews & F1 & Spearman & Accuracy & Accuracy & Accuracy & Accuracy\\
\midrule[0.5pt]		        
WikiBERT &1.01 & 1.98 &1.33 &2.43 &1.75 &7.01 & 52.32 & 15.03/15.02 \\
\model & 1.73 & 4.41 & 2.22 & 5.94 & 3.86 & 20.01 & 141.15 & 34.60/34.58\\
        \bottomrule[1.5pt]
\end{tabular}
}
\caption{Testing Latency of WikiBERT and \model measured by seconds. All experiments are computed in the same computational device with same batch size. The CPU is AMD EPYC 7K62 48-Core Processor. GPU is A100-SXM4. Driver Version is 450.156.00. CUDA Version is 11.1.}
\label{tab:pretrain_data}
\end{table*}

\section{Case Study}
We show three concrete examples from QNLI and ACL-ARC in Table~\ref{table:case_1},\ref{table:case_2},\ref{table:case_3}.
\label{appendix:case_study}

\section{More Experiments for Tuning \model}
In Table~\ref{table:more_tuning_experiments}, we show more results in Section~\ref{sec:tuning_model} on \texttt{STS-b}, \texttt{MRPC} and \texttt{QNLI}.
\label{appendix:tuning_model}
\begin{table}[H]
\centering
\resizebox{0.95\linewidth}{!}{
\begin{tabular}{ccccc}
\toprule[1.5pt]
      & \bf WikiBERT & \bf \model$_{\text{All}}$ &\bf  \model$_{\text{Fuse}}$ &\bf \model \\ \midrule[0.5pt]
STS-B & 88.64    & 86.82      & 89.20       & 89.10  \\ \midrule[0.5pt]
MRPC  & 88.85    & 87.42      & 91.27       & 91.54  \\ \midrule[0.5pt]
QNLI  & 90.66    & 88.19      & 91.36       & 91.28  \\ \bottomrule[1.5pt]
\end{tabular}}
\caption{Experimental Results as in Section~\ref{sec:tuning_model} on \texttt{STS-b}, \texttt{MRPC} and \texttt{QNLI}. The evaluation metrics are Spearman correlation, F1 score and Accuracy respectively.}
\label{table:more_tuning_experiments}
\end{table}

\section{Details for Wikipedia and Pubmed}
The source and size of Wikipedia and Pubmed are shown in Table~\ref{tab:domain_datasets}. And hyper-parameters for WikiBERT and PubmedBERT pre-training is shown in Table~\ref{table:wikibert_details}.

\label{appendix:pretraining_data}
\begin{table*}[htbp]
\centering
\small
\begin{tabular}{llll}
\toprule
\bf Dataset & \bf Domain & \bf Source & \bf Size \\
\midrule
Wikipedia
& General
& \url{https://dumps.wikimedia.org}
& 14.35GB\\
PubMed
& Biomedical
& \url{https://github.com/naver/biobert-pretrained}
&28.12GB\\
\bottomrule
\end{tabular}
\caption{List of the PubMed and Wikipedia.}
\label{tab:domain_datasets}
\end{table*}

\begin{table}[H]
    \centering
    \resizebox{0.9\linewidth}{!}{
    
    \begin{tabular}{cc}
        \toprule
        \textbf{Hyperparameter} & \textbf{Assignment}  \\
        \midrule
        vocab size & 30522 \\
        Uncased & True \\
        number of Layers & 12 \\
        attention Head & 12 \\
        ffn Size & 3072 \\
        mlm masking & static \\
        batch size & 64 \\
        gradient accumulation steps & 128 \\
        max train steps & 8000 \\
        optimizer &  FusedLAMBAMP \\
        learning rate & 6e-3 \\
        index refreshing step & 200\\
        learning rate scheduler & PolyWarmUpScheduler \\
        Warmup proportion & 0.2843 \\
        weight decay & 0.01 \\
        \bottomrule
    \end{tabular}}
    
    \caption{Hyperparameters for WikiBERT and PubmedBERT pretraining.} 
    \label{table:wikibert_details}
\end{table}

\begin{table*}[htbp]
\centering	
\resizebox{1\linewidth}{!}{
\begin{tabular}
    {m{2cm}m{10.5cm}cc}
\toprule[1.5pt]
\bf Question & \bf Answer  & \bf Prediction & \bf Label  \\
\midrule[0.5pt]
How much of Jacksonville is made up of water? & According to the United States Census Bureau, the city has a total area of 874.3 square miles (2,264 $\text{km}^2$), making Jacksonville the largest city in land area in the contiguous United States; of this, 86.66\% (757.7 sq mi or 1,962 $\text{km}^2$) is land and ; 13.34\% (116.7 sq mi or 302 $\text{km}^2$) is water.  & Entailment & Entailment   \\
\midrule[0.5pt]	
\addlinespace[0.15em]
\multirow{5}{*}{\bf Knowledge}
&\multicolumn{3}{|m{14.5cm}}{
(1)~this article lists the 3, 143 states of america. the 50 states of the united states are divided into 3, 007 " counties ", political and geographic subdivisions of a state ; 236 other local governments and geographic places are also first - order administrative divisions of their respective state / district / territory, but are called by different names. the latter are referred to collectively as " county equivalents " by the united states census bureau. the 236 county equivalents include 100 equivalents in the territories ( such as those in puerto rico ) outside the 50 states and the district of columbia. the large majority of counties and equivalents were organized by 1970. since that time, most creations, boundary changes and dissolutions have occurred in alaska and virginia. among the 50 states, 44 are partitioned entirely into counties, with no county equivalents. louisiana is instead divided into 64 equivalent parishes.} \\
&\multicolumn{3}{|m{14.5cm}}{
(2)~the united states census bureau ( usc \#\#b ) , officially the bureau of the census , is a principal agency of the u . s . federal statistical system , responsible for producing data about the american people and economy . the census bureau is part of the u . s . department of commerce and its director is appointed by the president of the united states . the census bureau ' s primary mission is conducting the u . s . census every ten years , which all \#\#oca \#\#tes the seats of the u . s . house of representatives to the states based on their population . [ 1 ] the bureau ' s various census \#\#es and surveys help all \#\#oca \#\#te over \$ 67 \#\#5 billion in federal funds every year and it assists states , local communities , and businesses make informed decisions . [ 2 ] [ 3 ] [ 4 ] the information provided by the census informs decisions on where to build and maintain schools , hospitals , transportation infrastructure , and police and fire departments}\\
&\multicolumn{3}{|m{14.5cm}}{
(3)~the crestview – fort walton beach – destin, florida, metropolitan statistical area, as defined by the united states census bureau, is a metropolitan area consisting of two counties in northwest florida, anchored by the cities of crestview, florida, and fort walton beach, florida. as of the 2010 census, the msa had a population of 235, 865, and a 2012 population estimate of 247, 665. the metropolitan area is a part of the " northwest corridor " which includes the pensacola metropolitan area and the panama city metropolitan area. demographics. as of the census of 2010, there were 235, 865 people, 95, 892 households, and 63, 964 families residing within the msa. the racial makeup of the msa was 81. 1 \% white, 9. 3 \% african american, 0. 3 \% native american, 2. 9 \% asian, 0. 1 \% pacific islander, 0. 2 \% from other races, and 3. 9 \% from two or more races. hispanic or latino of any race were 6. 8 \% of the population. according to the 2010 american community survey 1 - year} \\
&\multicolumn{3}{|m{14.5cm}}{
(4)~analog to digital conversions were achieved through steinberg, and in some cases mytek, converters. the album was recorded and mixed exclusively with steinberg cubase digital audio workstations on microsoft windows operating systems with waves ssl and abbey road tg12413 plugins. it was revealed that neither brahm nor marc know how to operate autotune, so it was not used. the songs were often performed to a click track, but there was no " snapping the drums to a grid ", which is a popular computerized technique to ensure that drums are in perfect time while simultaneously sucking the life out of an otherwise real performance. production. " tears of the enchanted mainframe " was produced and engineered by taylor and kaducak. backmasking is used on the track " superusurper " during an interlude that features a reversed reading of a passage from the george orwell novel nineteen eighty four. the album was mastered by geoff pesche and alex wharton at abbey road studios in london. title and artwork. " tears of the enchanted mainframe "} \\
&\multicolumn{3}{|m{14.5cm}}{
(5)~the zafarnama (, lit. " book of victory " ) is a biography of timur written by the historian nizam ad - din shami. it served as the basis for a later and better - known " zafarnama " by sharaf ad - din ali yazdi. one translation by felix tauer was published in prague in 1937. } \\

\bottomrule[1.5pt]
\end{tabular}}
\caption{Example from QNLI dataset.}
\label{table:case_1}
\end{table*}

\begin{table*}[h!]
\centering	
\resizebox{1\linewidth}{!}{
\begin{tabular}
    {m{2cm}m{10.5cm}cc}
\toprule[1.5pt]
\multicolumn{2}{c}{\bf Input}&  \bf Prediction & \bf Label  \\
\midrule[0.5pt]
\multicolumn{2}{m{10.5cm}}{Various approaches for computing semantic relatedness of words or concepts have been proposed , e.g. dictionary-based ( Lesk , 1986 ) , ontology-based ( Wu and Palmer , 1994 ; Leacock and Chodorow , 1998 ) , information-based ( Resnik , 1995 ; Jiang and Conrath , 1997 ) or distributional ( Weeds and Weir , 2005 ).} & Background & Background   \\
\midrule[0.5pt]	
\addlinespace[0.15em]
\multirow{5}{*}{\bf Knowledge}
&\multicolumn{3}{|m{14.5cm}}{
(1)~instrumentation and control engineering ( ice ) is a branch of engineering that studies the measurement and control of process variables, and the design and implementation of systems that incorporate them. process variables include pressure, temperature, humidity, flow, ph, force and speed. ice combines two branches of engineering. instrumentation engineering is the science of the measurement and control of process variables within a production or manufacturing area. meanwhile, control engineering, also called control systems engineering, is the engineering discipline that applies control theory to design systems with desired behaviors. control engineers are responsible for the research, design, and development of control devices and systems, typically in manufacturing facilities and process plants. control methods employ sensors to measure the output variable of the device and provide feedback to the controller so that it can make corrections toward desired performance. automatic control manages a device without the need of human inputs for correction, such as cruise control for regulating a car's speed. control systems engineering activities are multi - disciplinary in nature. they focus on the implementation of control systems, mainly derived by mathematical modeling. because instrumentation and control play a significant role in gathering information from a system and changing its parameters, they are a key part of control loops. as profession. high demand for engineering professionals is found in fields associated with process automation. specializations include industrial instrumentation, system dynamics, process control, and control systems. additionally, technological knowledge, particularly in computer systems, is essential to the job of} \\
&\multicolumn{3}{|m{14.5cm}}{
(2)~instrumentation is the art and science of measurement and control. instrumentation may also refer to:}\\
&\multicolumn{3}{|m{14.5cm}}{
(3)~the scientific and technological innovation ability of colleges and universities, and strengthening the evaluation research of the scientific and technological innovation ability and efficiency of colleges and universities, can we better promote the scientific and technological innovation ability of colleges and universities. universities the evaluation of scientific and technological innovation ability in colleges and universities is a complex system engineering, and the understanding of its connotation is the most important problem to be considered in the comprehensive evaluation. by consulting the data, it is found that the previous researches are mainly focused on the following three aspects : 1. from the perspective of innovative resource demand and innovative achievements, the scientific and technological innovation in colleges and universities is regarded as an organic whole composed of various elements. in the whole innovation system, colleges and universities undertake the functions and tasks of knowledge production and dissemination, technological innovation and transformation as well as personnel training. according to the relationship between innovation elements, the scientific and technological innovation ability of colleges and universities is divided into basic strength of scientific and technological innovation, scientific and technological innovation input ability, knowledge innovation ability, technological innovation ability, scientific and technological innovation output ability. science and technology innovation achievement transformation ability, talent innovation ability. 2. from the perspective of innovation process, the ability of scientific and technological innovation in colleges and universities is embodied in the process of knowledge creation, knowledge dissemination, transformation and diffusion of technological inventions. it also includes the technological, economic and managerial abilities that the university relies on} \\
&\multicolumn{3}{|m{14.5cm}}{
(4)~automation engineering has two different meanings : automation engineer. automation engineers are experts who have the knowledge and ability to design, create, develop and manage machines and systems, for example, factory automation, process automation and } \\
&\multicolumn{3}{|m{14.5cm}}{
(5)~this learning methodology is called blended learning. blended learning can also incorporate machine learning and other such technologies to implement adaptive learning.} \\

\bottomrule[1.5pt]
\end{tabular}}
\caption{Example from ACL-ARC dataset.}
\label{table:case_2}
\end{table*}

\begin{table*}[h!]
\small
\centering	
\resizebox{1\linewidth}{!}{
\begin{tabular}
    {m{2cm}m{10.5cm}cc}
\toprule[1.5pt]
\multicolumn{2}{c}{\bf Input}&  \bf Prediction & \bf Label  \\
\midrule[0.5pt]
\multicolumn{2}{m{10.5cm}}{Although there are other discussions of the paragraph as a central element of discourse ( e.g. Chafe 1979 , Halliday and Hasan 1976 , Longacre 1979 , Haberlandt et al. 1980 ) , all of them share a certain limitation in their formal techniques for analyzing paragraph structure .} & CompareOrContrast & CompareOrContrast   \\
\midrule[0.5pt]	
\addlinespace[0.15em]
\multirow{5}{*}{\bf Knowledge}
&\multicolumn{3}{|m{14.5cm}}{
(1)~automation engineering has two different meanings : automation engineer. automation engineers are experts who have the knowledge and ability to design, create, develop and manage machines and systems, for example, factory automation, process automation and warehouse automation. scope. automation engineering is the integration of standard engineering fields. automatic control of various control system for operating various systems or machines to reduce human efforts \& amp ; time to increase accuracy. automation engineers design and service electromechanical devices and systems to high - speed robotics and programmable logic controllers ( plcs ). work and career after graduation. graduates can work for both government and private sector entities such as industrial production, companies that create and use automation systems, for example paper industry, automotive industry, food and agricultural industry, water treatment, and oil \& amp ; gas sector such as refineries, power plants. job description. automation engineers can design, program, simulate and test automated machinery and processes, and usually are employed in industries such as the energy sector in plants, car manufacturing facilities or food processing plants and robots. automation engineers are responsible for creating detailed design specifications and other documents, developing automation based on specific requirements for the process involved, and conforming to international standards like iec - 61508, local standards, and other process specific guidelines and specifications, simulate, test and commission electronic equipment for automation.} \\
&\multicolumn{3}{|m{14.5cm}}{
(2)~abstract. manipulator is a powerful tool which can help people to carry out the safe operation, production automation and improve the productivity of labor. based on the summary of the situation of research and development of manipulator, this article analyzes the functions of parts moving manipulator and carries out mechatronic design of parts moving manipulator according to the practical project items of parts moving manipulator of enterprises. on the basis of the analysis of the performance requirement and the operating characteristics of parts moving manipulator, this article analyses and designs the whole schemes for the mechanical structure, driving system, driving mode and the software and hardware control system of manipulator, and in which, the form of mechanical structure of cylindrical coordinate system is determined to be adopted in the design of manipulator, the driving scheme of pneumatic transmission is adopted, and the system control is carried out by plc. on this basis, this article analyses the kinematics and dynamics of parts moving manipulator and summarizes the relationship between displacement, speed, acceleration and joint angle. with the progress of science and technology and the development of social economy, the application area of manipulator has been becoming wider and wide. the manipulator can be found everywhere in human society. the application of manipulator has been extended to the civilian application fields such}\\
&\multicolumn{3}{|m{14.5cm}}{
(3)~in working environments with large manipulators, accidental collisions can cause severe personal injuries and can seriously damage manipulators, necessitating the development of an emergency stop algorithm to prevent such occurrences. in this paper, we propose an emergency stop system for the efficient and safe operation of a manipulator by applying an intelligent emergency stop algorithm. our proposed intelligent algorithm considers the direction of motion of the manipulator. in addition, using a new regression method, the algorithm includes a decision step that determines whether a detected object is a collision - causing obstacle or a part of the manipulator. we apply our emergency stop system to a two - link manipulator and assess the performance of our intelligent emergency stop algorithm as compared with other models. increasing the safety of robots, especially industrial manipulators, is just as important as improving their performance. a collision between a manipulator and a person, for example, may cause severe personal injury as well as damage to the machinery. thus, it is necessary to develop an algorithm that can detect collisions before they occur and make the manipulator stop before damage is done. various emergency stop or obstacle avoidance algorithms for robots, particularly those utilizing distance - measuring sensors [ 1 ] [ 2 ] [ 3 ] [ 4 ] or vision sensors have been reported [ 5 ] [ 6 ] [ 7 ] [ 8 ] and those algorithms using each} \\
&\multicolumn{3}{|m{14.5cm}}{
(4)~the reliability of kinematic trajectory of manipulators describes the ability that manipulators keep kinematic accurate. it is an important parameter to evaluate the performance of manipulators. the kinematic accuracy of manipulators can be improved when piezoelectricity material are used as a transducer to suppress the vibration of flexible manipulators. first, a 3 degree - of - freedom parallel manipulator system and its dynamic equations are introduced. the theory and experiment of a vibration suppression system are then presented. the calculation method of both error and reliability of kinematic trajectory of manipulator is further implemented. finally, the reliability of kinematic accuracy are calculated and analyzed for the 3 degree - of - freedom parallel manipulator with or without vibration suppressing control. the results show that the reliability of kinematic accuracy is improved using vibration suppressing control. the reliability of kinematic accuracy of manipulators is an important indicator to evaluate the accuracy of manipulator motion [ 1 ]. in manipulators, light weight linkages are employed to achieve high speed and acceleration motions for better performance. however, the light weight linkage will result in inherent structural vibration, and the structural vibration leads to inaccurate kinematic trajectory of manipulators. different methods have been proposed to reduce the vibration of the flexible link} \\
&\multicolumn{3}{|m{14.5cm}}{
(5)~abstract - economic dispatch and frequency regulation are typically viewed as fundamentally different problems in power systems and, hence, are typically studied separately. in this paper, we frame and study a joint problem that co - optimizes both slow timescale economic dispatch resources and fast timescale frequency regulation resources. we show how the joint problem can be decomposed without loss of optimality into slow and fast timescale subproblems that have appealing interpretations as the economic dispatch and frequency regulation problems, respectively. we solve the fast timescale subproblem using a distributed frequency control algorithm that preserves network stability during transients. we solve the slow timescale subproblem using an efficient market mechanism that coordinates with the fast timescale subproblem. we investigate the performance of our approach on the ieee 24 - bus reliability test system. abstract - economic dispatch and frequency regulation are typically viewed as fundamentally different problems in power systems and, hence, are typically studied separately. in this paper, we frame and study a joint problem that co - optimizes both slow timescale economic dispatch resources and fast timescale frequency regulation resources. we show how the joint problem can be decomposed without loss of optimality into slow and fast timescale subproblems that have appealing interpretations as the economic dispatch and frequency regulation problems, respectively. we solve the fast timescale subproblem} \\

\bottomrule[1.5pt]
\end{tabular}}
\caption{Example from ACL-ARC dataset.}
\label{table:case_3}
\end{table*}

\end{document}